\documentclass{article}

\usepackage{arxiv}

\usepackage[utf8]{inputenc} % allow utf-8 input
\usepackage[T1]{fontenc}    % use 8-bit T1 fonts
\usepackage{hyperref}       % hyperlinks
\usepackage{url}            % simple URL typesetting
\usepackage{booktabs}       % professional-quality tables
\usepackage{amsfonts}       % blackboard math symbols
\usepackage{nicefrac}       % compact symbols for 1/2, etc.
\usepackage{microtype}      % microtypography
\usepackage{lipsum}
\usepackage{graphicx}
\graphicspath{ {./images/} }

\title{Training with reduced precision of a support vector machine model for text classification}

\author{
  Dominik Zurek \\
  Department of Computer Science\\
  AGH University of Science and Technology\\
  Cracow, Poland \\
  \texttt{dzurek@agh.edu.pl} \\
   \And
 Marcin Pietron \\
  Department of of Computer Science\\
  AGH University of Science and Technology\\
  Cracow, Poland \\
  \texttt{pietron@agh.edu.pl} \\
}

\begin{document}
\maketitle

\begin{abstract}
This paper presents the impact of using quantization on the efficiency of multi-class text classification in the training process of a support vector machine (SVM). This work is focused on comparing the efficiency of SVM model trained using reduced precision with its original form. The main advantage of using quantization is decrease in computation time and in memory footprint on the dedicated hardware platform which supports low precision computation like GPU (16-bit) or FPGA (any bit-width). The paper presents the impact of a precision reduction of the SVM training process on text classification accuracy. The implementation of the CPU was performed using the OpenMP library. Additionally, the results of the implementation of the GPU using double, single and half precision are presented.

\end{abstract}

\section{Introduction}
The classification problem is one of the most common tasks in an artificial intelligence. 
%Those group of algorithms belong to supervised learning where algorithms are learned from labeled data and then determine label for new input unlabelled example or unsupervised learning where algorithm has only input data and must find structure in its input.  
The most common classification algorithm includes logistic regression, k-NN, naive Bayes, decision trees, neutral networks(NN) and deep neutral networks (DL) etc. In the case of non-complex data boosted or bagged weak learners such as logistic regression and k-NN can achieve results close to SVM or deep learning models. Text classification is a task in which SVM is slightly inferior to CNN on some chosen text data benchmarks
\cite{svm_text_classification}\cite{svm_text1}\cite{svm_text3}.
%\textcolor{red}{i chyba deep neural networks} \textcolor{blue}{poprawione} or as it is presented in this paper Support Vector Machine(SVM). \textcolor{red}{nie rozumiem tego zdania} \textcolor{blue}{poprawione} 
One of the main leading features of SVM is its capability to learn input feature space and build a hyper-plane based on a margin that separates the data, rather than being based on number of features. The SVM very often requires less training data than NN to achieve an effective level of performance. In addition SVM are less prone to overfitting than NN \cite{overfitting_svm_cnn}. 
The SVM algorithm is suitable for text classification \cite{svm_text_classification} as an example of linearly separable data(see section \ref{linear svm}). Using this type of algorithm there is no necessity to apply \textit{the kernel trick} (see section \ref{non-linear svm}) which is time consuming operation(e.g. in case of the Gaussian kernel, the exponential function is applied).

In the case of machine/deep learning algorithms, the training process is very time and memory consuming, which is of huge significance when it comes to  hardware  implementation (FPGA or  GPGPU).
%In general a computing systems doing computation and storing data in their 64 or 32 bit floating-point representation. 
One technique to achieve better performance and save memory is bit-width reduction of data. This operation is known as \textit{quantization}. In our experiments, data are quantized to use their low-accuracy representation during the training process of SVM. SVMs trainded in such way along with the bias parameter are used for classification in which test data are quantized to the same precision as was the case in training. The main goal of this paper is 
%to show how using 
to show how reduced representation of data 
%can improve 
in the training process of the SVM affects final accuracy of classification text data.
%with no loss of efficiency with regard to classification of text data.

\section{Related works}
Support vector machines (SVM) have been applied to text classification \cite{svm_text_classification}\cite{svm_text1}\cite{svm_text3}, where the efficiency is between 80-96\% and is dependent upon the benchmark. SVM achieves results that are approximately 5\% worst than those attained through deep learning models \cite{svm_vs_cnn} in text classification. 
The best performance in many natural language processing tasks can be obtained using recurrent and convolutional neutral networks \cite{cnn_rnn_text}. \\
Recurrent neutral networks (RNN) and convolutional neutral network(CNN) depend on architecture which might include megabytes of coefficients. In order to compression RNN’s and CNN’s model weights pruning and weights quantization are applied. The effects of those approaches for RNN  \cite{rnn_fpga_quant}\cite{rnn_quant_2} and for CNN \cite{cnn_quantization_1}\cite{cnn_qunatiozation_2} has been presented in previous research. In both cases (RNN and CNN) the authors proposed the \textit{dynamic fixed point quantization} and the \textit{Integer quantization} methods to quantize the coefficient of the learned neural network and with that the parameters performing the classification. In some cases, retraining is performed after quantization. In this approach, to perform proper classification, it is sufficient use 5-bits in order to not lose efficiency. Another approach with regard to compressing a neutral network is applying quantization during the training phase. In this case, bit-widths of activation, gradients and weights are reduced. The training of RNN and CNN with reduced precision is shown in previous research  \cite{rnn_traning_quant_1}\cite{rnn_traning_quant_2} and \cite{cnn_traning_quant_1}, where the authors have shown that when it comes to classification problem, 6 bit-width must be used as a minimum to not cause a significant drop in accuracy. The quantization process has been applied on SVM algorithm and the results are presented in literature \cite{SvmQuant}.
\section{Natural language processing}
\label{nlp}
Natural language processing (NLP) is a branch of artificial intelligence which is focused on enabling an application to understand and process text or speech in a human language. NLP models are able to perform language translation, text classification, text categorization, similarity calculation, spelling correction, word sentence disambiguation, information extraction, and semantic understanding \cite{text_categorization}\cite{document_similarity}\cite{text_classification_with_vector_reduced_precision}\cite{cnn_quantization_1}\cite{sentence_classification} etc. 

%Those abilities are applied in the common applications like: chatbots, check grammatical, call centers and eCommerce.  

%In mobile devices era very popular are become application related to speech recognition and speech synthesis(transformation speech to text and vice versa). 
In our research, we are focused on text document classification. Deep learning algorithms are incapable of processing text in their raw form - they are designed to learn from numerical data.
%Most commonly used techniques to predict word base on its context use a conditional probability P(w|c). 
The most widely used methods (e.g. those used in this paper) in which numerical representation of words or terms determining their importance to a given document in corpora (the weighting of words/terms), is term frequencies (TF) and inverted document frequencies (IDF) \cite{tf_idf}. TF-IDF is an information retrieval technique in which the weight is calculated by two components: term frequency (TF) and inverse document frequency (IDF). The weights are calculated by: 
\begin{equation}
    w_{ij} = tf_{ij}\times log\frac{N}{n_{j}}
\end{equation}
where tfij is the frequency of term j in document i, N is the total number of documents, and nj is the number of documents containing term j. As a result, each word or term has its respective TF and IDF score, which is zero for common words. Another popular technique for conducting such mapping is word embedding. Word embedding refers to distributed representations of text with low-dimensional vectors, where words with similar semantics are mathematically closer, for example, ‘author’ and ‘paper’ are closer to each other than ‘author’ and ‘square’. The training method for a word embedding model is the relationship   between target $\textit{w}$ and its context \textit{c}. The most popular word embedding methods are \textit{word2Vec} \cite{Word2Vec} and \textit{GloVe} \cite{gloVec}. Both GloVe and Word2Vec are context independent - there is only one numerical representation of a word regardless of the position of the word in the sentence. Recent devised models \cite{BERT}\cite{Tranformers}  build with positional encoding to represent word position and in which different vectors are generated for a word in order to capture its context. As a result of the use of the above-mentioned methods, the performance of crucial NLP problems such as language modeling, question answering, document clustering and LSA (latent semantic analysis), have been significantly improved. In this paper, the TF-IDF method is used. This approach is one of the simplest, but it is suitable for text classification using SVM, because text is an example of a line separable task.
\section{Overview of Support Vector Machines}
\label{svm oerviev}
Support vector machines (SVM) are binary classifiers (there are exactly two classes of object) which for given labeled training data (supervised learning) create an optimal hyperplane in N-dimensional space (N is the number of features) which categorizes a set of positive examples from a set of negative examples with a maximum margin (distance of the hyperplane from the nearest of the positive and negative examples). The points closest to the separating hyperplane are known as support vectors (SVs) \cite{vapnik}. 
The main usage of SVM is image segmentation \cite{svm_image_1}\cite{svm_image_2}\cite{svm_image_3} and text classification \cite{svm_text_classification}\cite{svm_text1}\cite{svm_text3}.

\subsection{Theoretical background}
For a given training data set of n features with the form ($x_{1}$, $y_{1}$),\dots,($x_{n}$, $y_{n}$) where $y_{i}$ are either 1 or -1 is the label of point $\overrightarrow{x_{i}}$. Vectors with the class label +1 are the positive class and those with -1 are the negative class.

\subsection{Linear SVM}
\label{linear svm}

The linear SVM separates data in n-dimensional space with the use of the decision hyperplane defined as: 
\begin{equation}
   f(x) = w^Tx+b
\end{equation}
where $w \in \Re^n$ is the normal vector to the hyperplane and \textit{x} is the input vector. In linear case data are  separated by two parallel hyperplanes upon which points with the smallest margin(SVs) are placed.
The goal is to maximize the separating margin so the value of $\| w \| = \sqrt{w^Tw}$ should be minimized what is equivalent with minimization of:
\begin{equation}
        \min_{w,b} = \frac{\| w \|^2}{2}
\end{equation}
This problem is resolved using Lagrange multipliers defied as $\alpha$, with respect to Karush-Kuhn-Tucker(KKT) conditions \cite{kkt}. Finally, the decision function becomes: 
\begin{equation}
\label{svm decission}
f(a) = sgn(\sum_{i=1}^{n}\alpha_{i}y_{i}x_{i}^Ta + b)
\end{equation}

\subsection{Non-linear SVMs}
\label{non-linear svm}
SVMs can be generalized to a non-linear classifier. In order to achieve this, \textit{The kernel trick} is used to obtain non-linear hyperplane. The mapping from two-dimensional space into a higher-dimensional space is achieved by $\phi$ transformation  $\overrightarrow{x_{i}}$ -> $\phi( \overrightarrow{x_{i}}$). The dot product then becomes:
\begin{equation}
K(x_{i}, x_{j}) = \phi(\overrightarrow{x_{i}})^T\phi(\overrightarrow{x_{j}})
\end{equation}
where \textit{K} is a kernel function $\Re^n$x$\Re^n$ -> $\Re$. The most common form of radial basis function is a Gaussian distribution, calculated as:
\begin{equation}
K(x_{i}, x_{j})  = exp(-\gamma\| x_{i} - x_{j} \|^2)
\end{equation}

Now the decision function is: 
\begin{equation}
\label{kernel decision}
f(a) = sgn(\sum_{i=1}^{n}\alpha_{i}y_{i}K(x_{i}^Ta) + b)
\end{equation}

\subsection{Training of SVM}
\label{traning description}
The main goal in the training process is finding the pairs of alphas with positive values that meet KKT conditions. This is known as the large quadratic programming optimization problem (QP) which can be resolved by complex quadratic solvers which optimize a quadratic function of several variables. John Platt \cite{platt} introduced the sequential minimal optimization (SMO) algorithm. SMO is an iterative algorithm which breaks down the large optimization problem into many smaller problems and solves them analytically. Each of the smallest sub-problems of QP involves two alphas. At every step, two alphas are chosen and then the algorithm attempts to optimize these multipliers and updates the SVM with new optimal values.To optimize of finding the pairs of  $\alpha_{i}$ and  $\alpha_{j}$ Platt\cite{platt} proposed using a different heuristic method for each. In Platt’s proposal, two loops are utilized: : the outer loop for finding  $\alpha_{j}$ and the inner loop seeks $\alpha_{j}$ for the chosen $\alpha_{i}$. Firstly, the outer loop reviews all of the training examples and chooses those which do not satisfy KKT \cite{kkt} conditions and marks them as candidates for optimization. Subsequently, the outer loop iterates over the non-bound examples - multipliers that are on neither the upper nor lower boundary - and checks the same point; if for concrete examples the KKT conditions are violated, it is eligible for optimization. Iterating over non-bounded examples continues until all of the non-bound examples start to obey the KKT conditions within the error defined as  $\xi$. The algorithm then begins the next iteration over the entire training set. In the outer loop, two iterations occur alternately: single over the entire training set and multiple over the non-bound set. The process continues until the entire training set obeys the KKT conditions with respect to $\xi$. For efficient working there occur the maintaining and updating of a cache for storing the error for the \textit{ith} non-bound training sample defined as: 
\begin{equation}
    E_{i}=\mu_{i} - y_{i}
\end{equation}

When $\alpha_{i}$ is chosen, then the algorithm is searching the  $\alpha_{j}$ value. The index \textit{j} is chosen to maximize the value $|E_{i} - E_{j}|$. Depending upon the sign of $E_{i}$, the algorithm chooses the example with the minimum ($E_{i}$ is positive) or the maximum error $E_{j}$ ($E_{i}$ is negative).  In the event of the above heuristic method not making sufficient progress, the iteration over the non-bound set is started and searches for an example which can make sufficient progress. If progress is still not achieved, the SMO reviewing the entire training set until an example which can make sufficient progress is found.

In addtition to finding the $\alpha$ values, training process determines the bias parameter \textit{b}. In the SMO algorithm, the \textit{b} value is updated after each step. Vapnik \cite{vapnik} suggested taking the average of \textit{b} values estimated from two support
vectors, each from a different class, int order to obtain a more accurate estimation.

\subsection{Multiclass-SVM}
\label{multiclass description}
As described, the SVM is a binary classifier (see section \ref{svm oerviev}) which means it is capable of interpreting the following definition describing concrete searching object. In order to build a multi-class SVM classifier \textbf{one-versus-all(OVA)} \cite{ova} technique is used. In this approach, the problem of classifying N classes is split into N binary problems. During the training process, the samples of \textit{n-th} class are labeled as positive (+1) and the rest of N-1 examples as negative (-1). As a result, \textit{N} SVM models are created. When testing, the classifier makes a decision using the following formula: 
\begin{equation}
\label{ova_decision_function}
    f_{ova}(x) = argmax_{i=1\dots N}f_{i}(x)
\end{equation}
where f(x) is defined depends on the bias of the kernel type by formula (\ref{svm decission} - linear) or (\ref{kernel decision} - non-linear).

\section{Quantization}

In most deep learning algorithms, it often occurs that the problem is computing time and memory occupancy. Commonly used techniques to manage this involves using low precision,  which means the conversion of data values from their N-bits to n-bits (N > n) representation. The reducing precision transformation is known as  \textit{quantization}. The incorporation of quantization to machine and deep learning  algorithms brings them to a reasonable size and provides them a high level of performance accuracy \cite{text_classification_with_vector_reduced_precision}\cite{rnn_fpga_quant}\cite{cnn_quantization_1}\cite{cnn_qunatiozation_2}\cite{rnn_traning_quant_1}\cite{rnn_traning_quant_2}\cite{cnn_traning_quant_1}.\\
In the most cases, the floating-point representation of numbers uses the \textit{IEEE-754} standard which has a sign(a bit to indicate whether the number is positive or negative - \textbf{S}), a mantissa(fractional part - \textbf{F}) and an exponent(integer part - \textbf{E}). With this standard, floating-point numbers are represented in the form:
\begin{equation}
    (-1)^S \times F \times 2^E
\end{equation}
The described standard is called floating-point representation because the values of the mantissa bits can be placed along with the decimal point, based on the exponent’s given value. This is in contrast to fixed-point representation, where the decimal point is always in the same place among the given bits. The IEEE-754 standard defines a number of different binary representations such as: half precision (E=5bits, F=10bits), single- precision(E=8bits, F=23bits), double-precision(E=11bits, F=52bits) and quadruple-precision(E=15bits, F=112bits).
In our experiments, a set of floating-point data is mapped to its reduced format using two different dynamic-fixed point quantization methods. Both of the proposed approaches are examples of linear quantization and they were used efficiently for qunatization CNN\cite{cnn_quantization_1} \cite{cnn_qunatiozation_2} and RNN\cite{rnn_fpga_quant} \cite{rnn_quant_2}. In this investigation we decided to conduct a SVM training process using both of them and to check the extent to which a drop in accuracy it would occur. In addition, fixed-point representation like \textit{fp16} is desired in modern GPU architecture to use special cores calling \textit{Tensor cores}\footnote{https://developer.nvidia.com/tensor-cores} which are intended to accelerate basic mathematical operations such as matrix-multiplication.

\subsection{Max magnitude dynamic fixed-point quantization}
\label{max fixed-point}

The first proposed quantization method to represent a floating-point value on its fixed-point \textit{$n_{total\_bits}$}-width representation needs to decide how many bits will be used to represent the integer \textit{$n_{integer\_part}$} and fractional part \textit{$n_{fractional\_part}$} part.
Determining the $n_{integer\_bits}$ for each concrete value is a realization based on its maximum possible absolute value which can occur during the entire SVM training process (e.g. when the max value is 3, there needs to be at least 2-bits used for the integer part to represent this). Statistics describing the maximum absolute input/output values for each sub-calculation of the training process are collected during pre-training experiments. Having this set, the number of integer bit is calculated by: 
\begin{equation}
    n_{integer\_bits} = ceil(\log_2(max|X|)
\end{equation}
and knowing this: 
\begin{equation}
    n_{fractional\_bits} = n_{total\_bits} - n_{integer\_bits} -1
\end{equation}
the quantization of the $\textit{X}$ value of operation \textit{p} can be realized by the following general formula: 

\begin{equation}
    q_{X_{p}} = 2^{-frac\_bits_{p}}\times round(2^{frac\_bits_{p}}\times X_{p})
\end{equation}

where the expression $2^{+/-(frac\_bits)}$ is the shifting of input \textbf{X} up or down. A drawback of the described approach may be loss of precision for large data distribution. Additionally \textit{representation saturation} can occur which means that integer bits cannot accommodate the full dynamic range of its original representation. As mentioned, before applying quantization, there is a running statistic gathering to avoid the unwanted appearance of. Building statistics is obviously run on a small subset of the original data-set, so for a small percentage of data, this side effect can occur.

\subsection{Min-max dynamic fixed-point quantization}
\label{min-max quantization}
A second quantization method for the given set of floating-point numbers takes as its parameters the minimum and maximum values \textit{$R_{min}$}, \textit{$R_{max}$} and \textit{bits-width} for the fixed-point representation of a floating-point. Having this, the i-th fixed-ranges for given set of floating-point values are built using the form : 
\begin{equation}
\label{fixed-range}
    range_{i} = \Bigg[r_{max_{i-1}}; r_{min_{i}} + \frac{R_{max} - R_{min}}{2^{Bitwidth}}\Bigg)
\end{equation}
where \textit{$r_{min_{i}}$} = \textit{$r_{max_{i-1}}$} and \textit{$r_{min_{0}}$} = \textit{$R_{min}$}.

The same occurred in the previous method (see section \ref{max fixed-point}) the possible minimum and maximum  input/output values for each sub-calculation of the SVM training process are gathered during pre-training experiments.\\ 
Each operation \textit{p} has its own set \textit{$S_{p}$} of ranges. The quantization of the \textit{$X_{p}$} value is realized by finding the range  \textit{$range_{p}$} from the set \textit{$S_{p}$} and replacing this value by the \textbf{median value} of \textit{$range_{p}$}.
In some special cases, this method  could lead to changing the sign of quantization value. This effect may significantly breakdown the efficiency of the algorithm. To avoid such a situation, there is a necessity to ensure that \textit{$sign(r_{min_{i}}$) = sign($r_{max_{i}}$)}. It is required to split the range which contains inside 0.f like: [\textit{$r_{min_{i}}$}\dots 0\dots \textit{$r_{max_{i}}$}) into two separate ranges like: [\textit{$r_{min_{i}}$}\dots 0) and (0\dots \textit{$r_{max_{i}}$}). Both new ranges are half open from 0.f side because 0.f should not be quantized as this could produce an unwanted effect.

\section{Implementation}
\label{implementation}
  As training and testing data, the following were used: the Reuters Dataset-r8(Reuters articles with single label from R10 sub-collection  of Reuters-21578)\footnote{https://martin-thoma.com/nlp-reuters/} and WebKB\footnote{http://www.cs.cmu.edu/afs/cs.cmu.edu/project/theo-20/www/data/} data-sets which are multi-class (i.e there are multiple classes) and multi-label (i.e. each document can belong to many classes) data sets where documents are grouped by class. The main goal of the training process is obtained by the numbers of support vector machines which are able to classify documents. As described,  the SVM is a binary classifier (see section \ref{svm oerviev}) ) so in order to realize multi-class classification, the OVA algorithm is used (see section \ref{multiclass description}). 
\subsection{Implementation steps}
The following modules were developed to calculate the SVs Vectors for one class:
\begin{enumerate}
    \item Statistic collection - this application was run before the training process and it was realized by gathering information concerning the maximum and minimum values for each training calculation. These values are used in the quantization process (see sections : V-A and V-B) \ref{min-max quantization}).
    \item TF-IDF - input text data are represented in their numerical form. In addition to this step, data are cleaned by: turning all the letters to lower case, removing stopwords, removing one- and two-letters words and stemming words (the process of reducing inflected words to their word stem, base or root form). 
    \item Calculating $\alpha$ and \textit{b} parameters (see section \ref{svm oerviev}) - this is the main part of the entire process. As input this module takes the training data in the form of a matrix and processes them according to the \textbf{SMO} algorithm, which is precisely described in section \ref{traning description}. During this calculation, there occurs the realization of time-consuming operation, such as matrix multiplication which is conducted in a parallel manner. Before and after each operations, the input parameters and its results are quantized using the previously described method. As a result, the module return a set of $\alpha$ and bias parameter \textit{b} which describe support vector machines.

\end{enumerate}

This process is repeated N-times where N is the number of classes (8 in R-8 data-set case and 4 in the WebKB data- set). As a result, N-sets of vectors are created which each collect vectors which are able to classify each incoming document to a concrete class. As a last step, incoming training and test documents are classified, on the basis of the decision function defined by formula  (\ref{ova_decision_function}).

\subsection{CPU implementation}
The CPU's implementation of the SVM training process (see section \ref{traning description}) was performed with the C++ language using the OpenMP\footnote{http://www.openmp.org/} library to improve performance. Thanks to OpenMP, the calculations are conducted in a parallel manner. In the experiments, a special compiler directive $\textit{-03}$ was used to optimize multi-core CPU implementation. The experiments  were performed on Intel(R) Xeon(R) CPU E5-2630 v3(2.4GHz).

The main goal of this implementation is precision reduction which changes the floating-point representation of a number to its fixed-point representation using the two methods described in sections   \ref{max fixed-point} and \ref{min-max quantization}. The main goal of this implementation is precision reduction which changes the floating-point representation of a number to its fixed-point representation using the two methods described in - calculations are executed by the same hardware. Therefore, this implementation plays the simulation role. By reducing precising, we can check what the minimum number bits is which must be used for conducting the SVM training process without a significant drop in accuracy. This information could be used in implementation on some special platform such as FPGA in which no standard data width nor format is defined resulting in improved performance.       

The results of the described implementation for the proposed quantization method (see sections \ref{max fixed-point}, \ref{min-max quantization}) using two data sets are shown in Tables \ref{tab:Magnitude_res} and \ref{tab:Min_max_res}. In both cases, the significant drop in accuracy occurs for 5-bit and 4-bit precision. The main cause of this for the \textit{Max magnitude dynamic fixed-point quantization method} (see section \ref{max fixed-point})  is the huge amount of zeros which are the results of this quantization process for less than 6-bit precision. In the case of \textit{Min-max dynamic fixed-point quantization} (see section \ref{min-max quantization}) this effect is caused by an insufficient number of produced buckets of less than 6 bit-width so as a result, in many cases quantization returns the same value for many numbers and consequently, there is not the possibility of performing a proper calculation.

\begin{table}[ht]
\centering
\begin{tabular}{|l|l|l|l|l|l|l|l|}
\hline
{\textbf{Datasets}} & \multicolumn{7}{l|}{\textbf{Accuracy{[}\%{]} for selected precision{[}bits{]}}}            \\ \cline{2-8} 
& \textbf{32} & \textbf{16} & \textbf{8} & \textbf{7} & \textbf{6} & \textbf{5} & \textbf{4} \\ \hline
Reuters r-8                        & 95.59       & 95,59       & 95,03  & 94,81 & 94,97 & 91,9  & 90,43 \\ \hline
WebKG & 82,82 &  82,77   &  81,94   &  81,21 &   81,32  &  73,25    &   70,34         \\ \hline
\end{tabular}
\caption{Max magnitude dynamic fixed-point quantization results}
\label{tab:Magnitude_res} 
\end{table}

\begin{table}[ht]
\centering
\begin{tabular}{|l|l|l|l|l|l|l|l|}
\hline
{\textbf{Datasets}} & \multicolumn{7}{l|}{\textbf{Accuracy{[}\%{]} for selected precision{[}bits{]}}}            \\ \cline{2-8} 
                                  & \textbf{32} & \textbf{16} & \textbf{8} & \textbf{7} & \textbf{6} & \textbf{5} & \textbf{4} \\ \hline
Reuters r-8                        & 95.59       & 95,02       & 94,82      & 94,67      & 93,94     & 92,99      & 92,05      \\ \hline
WebKG  & 82,82       &      82,79      &    81,81        &    81,46        &     81,65       &     74,91       &   70,16         \\ \hline
\end{tabular}
\caption{Min-max dynamic fixed-point quantization results}
\label{tab:Min_max_res}
\end{table}

\subsection{GPGPU implementation}
The obtaining of the SVs process on GPGPU (general-purpose computing on graphics processing units)  platforms was realized wit the CUDA\footnote{https://developer.nvidia.com/cuda-gpus} language using CUBLAS\footnote{https://developer.nvidia.com/cublas} library. The calculations were performed on Nvidia Tesla V100-SXM2-32GB\footnote{https://www.nvidia.com/en-us/data-center/v100/}.  
The graphics processing units (GPUs) are in particular vvery effective for accelerating large matrix products such as Matrix-Vector multiplication, Vector-Vector multiplication and Dot-product of two vectors. These operations are at the heart of the SVM training process and in this case, are calculated on the GPU. These time-consuming operations could be improved by using  \textit{Tensor cores(TC)} which are a part the newest GPGPUs. In order to run calculation on TC, data must be represented by \textit{fp16, int8} or \textit{int4}. Using the quantization method proposed in this paper, it is possible to use only the \textit{fp16} data type. Due to the improvement outlined in the previous section, this representation will not cause a drop in accuracy (see table \ref{tab:Magnitude_res} and \ref{tab:Min_max_res}). Cuda-Math-Api \footnote{https://docs.nvidia.com/cuda/cuda-math-api/} is used in order to perform calculations with half the precision on the GPU. Cuda-Math-Api provides transformations and mathematical functions for half type. Besides the operations of the matrix in the training process, there is a necessary to conduct basic mathematical operations (e.g. calculating the bias value) on single numbers using the CPU.

The main purpose of this section is to present how using reduced precision on the GPU can influence the calculation time and memory footprint of the whole training process. 
The Reuters Dataset-r8(see seq.\ref{implementation}) was used as input for time performance measurement.  TF-IDF(see seq.\ref{nlp}) transformation generated a matrix with the shape (2538, 827). Table \ref{tab:gpu_time} contains time results for particular single matrix operations for the different data types by using the generated matrix. The results include data copying time from the processor to the GPU and vice-versa. The time needed to calculate the dot product is comparable with a less complex vector multiply vector operation because a single number is the result of this operation, thus the time necessary to copy data is not visible. Realizing all training with half precision has achieved a 1.26 x speed increase over single precision and 1.76 x over double precision. When employing the Reuter’s matrix on the GPU, 16.8 [MB] is occupied for double-precision. By using single and half precision, this number is decreased by two and four times, respectively.

\begin{table}[ht]
\centering
\begin{tabular}{|l|l|l|l|}
\hline
\textbf{Type} & \textbf{\begin{tabular}[c]{@{}l@{}}Vector*Vector\\ {[}ms{]}\end{tabular}} & \textbf{\begin{tabular}[c]{@{}l@{}}Matrix*Vector\\ {[}ms{]}\end{tabular}} & \textbf{\begin{tabular}[c]{@{}l@{}}Dot product\\ {[}ms{]}\end{tabular}} \\ \hline
double        & 0,66                                                                      & 3,04                                                                      & 0,64                                                                    \\ \hline
float         & 0,59                                                                      & 1,95                                                                      & 0,58                                                                    \\ \hline
half          & 0,52                                                                      & 1,12                                                                      & 0,51                                                                    \\ \hline
\end{tabular}
\caption{\label{tab:gpu_time}Time result of particular matrix operations}
\end{table}

\section{CONCLUSION AND FUTURE WORK}
This work is focused on a combination of NLP, SVM, document classification and precision reduction.We have presented the results for the quantization effect during the SVM training process on text classification. Additionally, we have examined the influence of conducting calculations using reduced precision on time efficiency and the memory footprint on the GPGPU. The results of the presented experiments prove that the proposed concept significantly speeds up computation and saves memory occupation on the hardware platform. Moreover there is exactly presented how improvement could be achieved by using reduced precision of data in popular matrix operations which are base on most deep learning algorithms which in consequence can optimize the time necessary to execute all algorithms. The authors have proposed two quantization methods which for precision levels higher than five bit do not cause significant loss of accuracy ($\sim1\%$) during the classification process. %This is one bit more than minimum number of bit which are necessary to realize classification process by CNN and RNN with accepted error.% 
This information could be used for implementation of SVM classifier on different hardware platform, such as FPGA or DSP processors which give discretion for quantization  adjustment. When it comes to GPGPU and CUDA frameworks, the presented method could be used in a conversion library like  Cuda-Math-Api which provides an interface to change data representation from double or float to half precision. We are not presenting execution time for CPU implementation, as a result changing the bit-width will be exactly the same for each bit-width. For the CPU's implementation, float type data is used but data are reduced to fixed-bit representation by applying two quantization methods which allow the performance of a simulation of the operation with reduced precision.
Modern GPGPUs accelerate mixed precision models by using \textit{Tensor cores}. The quantization methods proposed in this paper allows the use of only one type of data supported by TC, i.e. \textit{fp16}. Other types of data which could further improve calculation are  \textit{int8} and \textit{int4} so we can consider applying an integer quantization method for the SVM training process. An integer quanization method is used by the Tensorflow\footnote{tensorflow.org} framework and was successfully applied for CNN \cite{cnn_quantization_1} and RNN \cite{rnn_quant_2} quantization. The TF-IDF transformation generates a matrix with a sparsity $\sim90\%$  which is why future work also includes applying the training process using a sparse matrix approach as well. Moreover we consider using a non-linear quantization  method during the SVM training process which allows using less than 6-bit width, for the SVM training process.

\bibliographystyle{unsrt}  


\end{document}